\documentclass[10pt,twocolumn,letterpaper]{article}

\usepackage{cvpr}              % To produce the CAMERA-READY version
\definecolor{cvprblue}{rgb}{0.21,0.49,0.74}
\usepackage[pagebackref,breaklinks,colorlinks,allcolors=cvprblue]{hyperref}

\def\paperID{xxx} % Enter your 3DV Paper ID here once assigned
\def\confName{3DV\xspace}
\def\confYear{2027\xspace}

\title{Extrapolation Matters: Sparse-Only Depth Completion Beyond LiDAR Coverage}

\author{Maximilian Jansen\\
CARIAD SE\\
{\tt\small maximilian.jansen@cariad.technology}
\and
Gasser Elazab\\
CARIAD SE\\
{\tt\small gasser.elazab@cariad.technology}
\and
Praveen Kulkarni\\
Bosch\\
{\tt\small praveenanil.kulkarni@in.bosch.com}
}

\begin{document}
\twocolumn[{%
\renewcommand\twocolumn[1][]{#1}%
\maketitle
\begin{center}
    \centering
    \captionsetup{type=figure}
    \includegraphics[width=\textwidth]{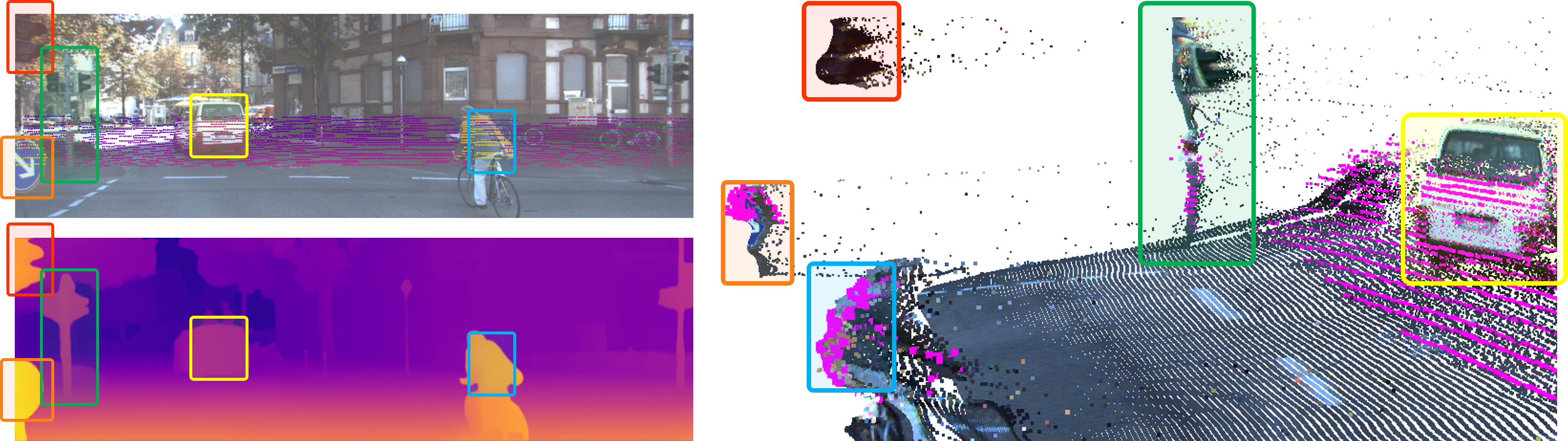}
    \caption{Qualitative example on the beyond-coverage extrapolation setting. \textbf{Left:} input RGB image with a sparse central LiDAR band and the corresponding predicted depth map. \textbf{Right:} colored 3D point cloud reconstructed from our prediction. The colored boxes indicate representative behaviors, including \textcolor{ForestGreen}{upward extrapolation of thin structures}, \textcolor{orange}{downward extrapolation}, \textcolor{black}{coherent surface reconstruction}, \textcolor{red}{plausible depth prediction for objects outside the LiDAR-supported region}, \textcolor{RoyalBlue}{suppression of spurious LiDAR measurements}, and \textcolor{violet}{consistency with valid LiDAR support}.}
    \label{fig:teaser}
\end{center}%
}]

\begin{abstract}
Extrapolation beyond LiDAR coverage is an important yet under-evaluated challenge of depth completion. In real camera--LiDAR systems, partial sensor overlap, occlusions, and limited field coverage often leave large image regions weakly observed or completely unobserved by LiDAR. Thus, we investigate depth completion in this beyond-coverage regime using only sparse supervision from raw single-sweep LiDAR, without dense depth labels, LiDAR accumulation, or pseudo-LiDAR generation. Our approach leverages a frozen monocular depth foundation model, lightweight per-frame alignment to sparse LiDAR geometry, and a refinement module that fuses the aligned prior with sparse LiDAR anchors to predict dense metric depth. To regularize learning under raw sparse supervision, we introduce a high-frequency normal-consistency loss defined as the image-space gradient of the per-pixel normal cosine discrepancy. This term penalizes spatial variation in scalar angular error; it does not directly match changes in neighboring normal vectors. We further introduce an evaluation protocol that explicitly separates LiDAR-supported and beyond-coverage regions, revealing extrapolative behavior that conventional depth-completion metrics can obscure. Our method achieves strong depth accuracy beyond LiDAR coverage while remaining competitive in LiDAR-covered regions, providing a practical way to deploy monocular foundation priors in multi-sensor systems.
\end{abstract}

\section{Introduction}
\label{sec:intro}

Depth completion estimates dense, metrically consistent depth from an RGB image and sparse range measurements. It is a core component of 3D perception in autonomous driving and robotics, where reconstruction, obstacle avoidance, and planning depend on reliable geometry~\cite{jeon2023mesh,maffra2019real}. However, in real automotive setups, LiDAR is both sparse and imperfect: single-sweep measurements are affected by scanline sparsity, multipath returns, calibration errors, and adverse weather, and large parts of the image may receive few or no returns at all~\cite{hu2022deepdepthcompletionextremely,kitti_data}. This makes extrapolation beyond LiDAR coverage, using only sparse and noisy supervision, an important but under-emphasized aspect of depth completion. 

Standard benchmarks are dominated by interpolation and refinement in LiDAR-supported regions, often using semi-dense targets obtained through multi-frame accumulation and stereo post-processing~\cite{elazab2026card,unilips,KITTI_improved_GT,yang2019drivingstereo}. In practice, camera-LiDAR systems frequently exhibit partial overlap, occlusions, and limited field coverage, leaving large image regions weakly observed or completely unobserved by LiDAR. Moreover, dense and clean ground-truth depth is often infeasible to obtain in real deployments, especially in sensor setups where stereo cameras cannot be mounted. A practical system must therefore do more than denoise and interpolate sparse depth: it must infer coherent geometry where no direct LiDAR support is available without relying on dense depth ground truth.

Monocular depth foundation models provide a natural prior for this setting, as they produce dense, highly generalizable depth predictions from a single image.~\cite{DepthAnything2024,bhat2023zoedepth,UniDepth2024,MoGe2025,MoGE_V2}, but their outputs are not directly usable as metric depth because they can suffer from scale, shift, and projection ambiguities~\cite{DepthAnything2024,MoGe2025}. The key question is therefore how to combine such priors with sparse LiDAR in a way that preserves their extrapolative strength while recovering metric consistency from raw sensor measurements.

In this work, we study exactly this problem in the sparse-only regime. Given a single RGB image and a raw LiDAR scan, we predict dense metric depth without dense depth labels, LiDAR accumulation, or pseudo-LiDAR generation. Our formulation combines a frozen monocular depth foundation model that provides a dense prior, a lightweight per-frame alignment of that prior to sparse LiDAR geometry, and a refinement module that fuses the aligned prior with sparse LiDAR anchors. The overall design is modular, so individual components can be replaced without changing the training recipe.

Learning in this regime is challenging because supervision is restricted to raw single-sweep LiDAR. Such supervision can contain localized spurious returns, and naive sparse-only training may propagate these artifacts into dense predictions. To address this, we combine robust sparse anchoring and surface-normal consistency with a high-frequency normal-consistency loss that suppresses localized geometric corruption while preserving low-frequency corrections to the dense prior.

Finally, we argue that standard depth-completion metrics are insufficient on their own, because metrics dominated by LiDAR-covered regions can obscure both behavior in LiDAR-unobserved regions and sensitivity to sparse measurement artifacts. We therefore introduce an extrapolation-focused evaluation protocol that explicitly separates LiDAR-supported and beyond-coverage regions, allowing us to distinguish interpolation of sparse anchors from extrapolation of coherent geometry. Hence, we summarize our contributions as follows:
% \begin{itemize}
    % \item \textbf{Depth completion beyond LiDAR coverage in the sparse-only regime.}
    % We argue that standard depth-completion evaluation under-emphasizes extrapolation, and we focus on the practically relevant setting where supervision is restricted to raw single-sweep LiDAR rather than dense labels, LiDAR accumulation, or pseudo-LiDAR generation.

    % \item \textbf{Sparse-only fusion combining aligned monocular priors with raw LiDAR.}
    % Our formulation uses a frozen monocular depth foundation model for dense structure, lightweight per-frame alignment for metric calibration, and a refinement module that fuses the aligned prior with sparse LiDAR anchors to predict dense metric depth.

    % \item \textbf{A high-frequency normal-consistency loss. }
    % By penalizing the image-space gradient of the per-pixel normal cosine discrepancy, the proposed objective reduces localized artifacts caused by spurious LiDAR returns and improves the robustness of sparse-only training.

    % \item \textbf{An evaluation protocol that separates interpolation from extrapolation.}
    % Our controlled protocol explicitly distinguishes LiDAR-supported from beyond-coverage regions, enabling a more faithful assessment of extrapolative depth completion. % and complementing conventional metrics that may not fully reflect robustness to sparse measurement artifacts.
% \end{itemize}

    \begin{itemize}

    \item \textbf{Sparse-only fusion of aligned monocular priors with raw LiDAR.}
    We propose a simple fusion formulation that aligns a frozen monocular depth foundation prior to sparse LiDAR geometry and refines it with a lightweight completion module to predict dense metric depth.

    \item \textbf{High-frequency regularization and beyond-coverage evaluation.}
    We introduce a high-frequency normal-consistency loss that suppresses localized artifacts under raw sparse supervision, and an evaluation protocol that separates LiDAR-supported and beyond-coverage regions to distinguish interpolation from extrapolation.

    \item \textbf{Sparse-only depth completion beyond LiDAR coverage.}
    We study depth completion in a practical setting where supervision is restricted to raw single-sweep LiDAR, and we highlight extrapolation beyond LiDAR coverage as a central but under-evaluated aspect of the task.
    
\end{itemize}

\section{Related Work}
\label{sec:related_work}

Depth completion has advanced through stronger completion architectures, richer supervision, and, more recently, monocular depth foundation models. However, most prior work is evaluated with metrics dominated by LiDAR-supported regions, where the task reduces to local interpolation. This leads to a metric saturation that obscures the model's ability to reason about global geometry in the extrapolative regime, where no direct LiDAR support exists.

\paragraph{Image-guided depth completion.}
Depth completion estimates dense metric depth from sparse range measurements, typically guided by an aligned RGB image. Early approaches relied on hand-crafted interpolation, sparsity-invariant convolutions, and edge-aware regularization to propagate sparse depth while respecting image structure~\cite{sparse_to_dense_2018,BilateralSolver_2016}. Modern methods instead learn this propagation directly, including guided and content-adaptive CNNs~\cite{guidenet,hu2021penet}, propagation-based methods such as CSPN~\cite{CSPN_compleition_2020}, NLSPN~\cite{NLSPN_2020}, and DySPN~\cite{DySPN_2023}, and architectures with explicit confidence handling or pre-propagation of sparse inputs~\cite{hu2021penet,BP-NET_2024}. Transformer-based and hybrid models such as GuideFormer~\cite{guideformer} and CompletionFormer~\cite{completionformer} further improve image-guided completion by combining long-range context with local refinement. Geometry-aware approaches additionally incorporate depth--normal constraints to better preserve local 3D structure~\cite{DeepLidar_2019_depthcompletion,xu2019depthnormalconstraints}. Our work is complementary to this line: rather than proposing a new completion architecture, we study how a completion backbone can be driven by a frozen monocular prior and raw sparse LiDAR under sparse-only supervision.

\paragraph{Supervision regimes for depth completion.}
Many depth completion methods are trained with densified supervision. On outdoor benchmarks such as KITTI~\cite{kitti_data,KITTI_improved_GT}, commonly used reference depth is obtained by accumulating multiple LiDAR sweeps and filtering them with stereo post-processing cues~\cite{KITTI_improved_GT}. As a result, high-performing methods are often trained on semi-dense labels rather than raw single-sweep LiDAR~\cite{NLSPN_2020,DySPN_2023,completionformer}. Alternative strategies reduce dependence on dense labels through photometric consistency or multi-view constraints~\cite{Ma2019SparseToDenseSelfSupervised,Monodepth2,wong2021unsupervised}. In contrast, we focus on a sparse-only setting aligned with deployment: training uses the same single-frame RGB image and raw sparse LiDAR measurements available at inference time, without densified labels or multi-frame supervision.

\paragraph{Monocular depth priors and foundation models.}
Recent monocular depth models provide increasingly strong dense geometric priors from a single image, including DepthAnything~\cite{DepthAnything2024,DepthAnythingV2_2025}, UniDepth~\cite{UniDepth2024,UniDepth2025_V2}, Metric3D~\cite{Metric3D_2023,Metric3D_v2}, and MoGe~\cite{MoGe2025,MoGE_V2}. A natural next step is to transfer such priors to depth completion, as demonstrated by DMD3C~\cite{DMD3c_2025}, which distills a monocular foundation model into a completion network via a two-stage pipeline: pseudo-LiDAR generation for large-scale pre-training followed by fine-tuning on real sparse LiDAR with a scale- and shift-invariant objective. Recent work also explores guided diffusion for zero-shot monocular depth completion and foundation-model-guided sparse-depth propagation for out-of-distribution completion~\cite{viola2025marigolddc,chen2025propagating}. Our approach targets a different operating point: rather than distilling or propagating the prior through a separate foundation-model pipeline, we align the frozen prior directly to sparse LiDAR geometry and fuse it with a completion backbone trained using sparse-only supervision from raw single-sweep measurements.

% Recent monocular depth models provide increasingly strong dense geometric priors from a single image, including DepthAnything~\cite{DepthAnything2024,DepthAnythingV2_2025}, UniDepth~\cite{UniDepth2024,UniDepth2025_V2}, Metric3D~\cite{Metric3D_2023,Metric3D_v2}, and MoGe~\cite{MoGe2025,MoGe_V2}. A natural next step is to transfer such priors to depth completion. The closest recent example is DMD3C~\cite{DMD3c_2025}, which distills a monocular foundation model into a completion network via a two-stage pipeline of pseudo-LiDAR pre-training and sparse LiDAR fine-tuning. This approach provides a strong baseline by leveraging the extensive geometric knowledge of foundation models; however, our approach targets a different operating point. Rather than distilling the monocular prior into a new network, we align the frozen prior directly to sparse LiDAR geometry and fuse it with a completion backbone trained using sparse-only supervision from raw single-sweep measurements.

\paragraph{Extrapolation beyond LiDAR coverage.}
Most prior work evaluates depth completion with metrics dominated by LiDAR-supported regions~\cite{DySPN_2023,DMD3c_2025,BP-NET_2024}, where the task is primarily interpolation, denoising, and local refinement. Yet real camera-LiDAR systems often exhibit partial overlap, occlusions, and persistent coverage gaps, making extrapolation beyond LiDAR coverage a practically important regime. Although methods based on strong monocular priors can exhibit extrapolative behavior qualitatively, this capability is rarely isolated or measured explicitly. Our work addresses this gap by studying sparse-only fusion with raw LiDAR and monocular foundation priors and by introducing an evaluation protocol that separates LiDAR-supported and beyond-coverage regions.

\section{Method}
\label{sec:method}

We propose a sparse-only fusion framework for depth completion beyond LiDAR coverage. The method combines a frozen monocular depth prior, a lightweight per-frame alignment to raw sparse LiDAR, and a fusion module that refines the aligned prior into dense metric depth. This decomposition separates \emph{global metric calibration} from \emph{local geometric refinement}: the monocular prior provides dense scene structure, alignment resolves global scale and shift ambiguity, and the fusion module incorporates sparse LiDAR measurements where available. To ensure robust training under raw sparse supervision, we apply geometric regularization via a high-frequency normal-consistency loss that suppresses localized artifacts from spurious LiDAR measurements while preserving large-scale structure.

\subsection{Pipeline Overview and Architecture}

\begin{figure*}[t]
    \centering
    \includegraphics[width=\linewidth]{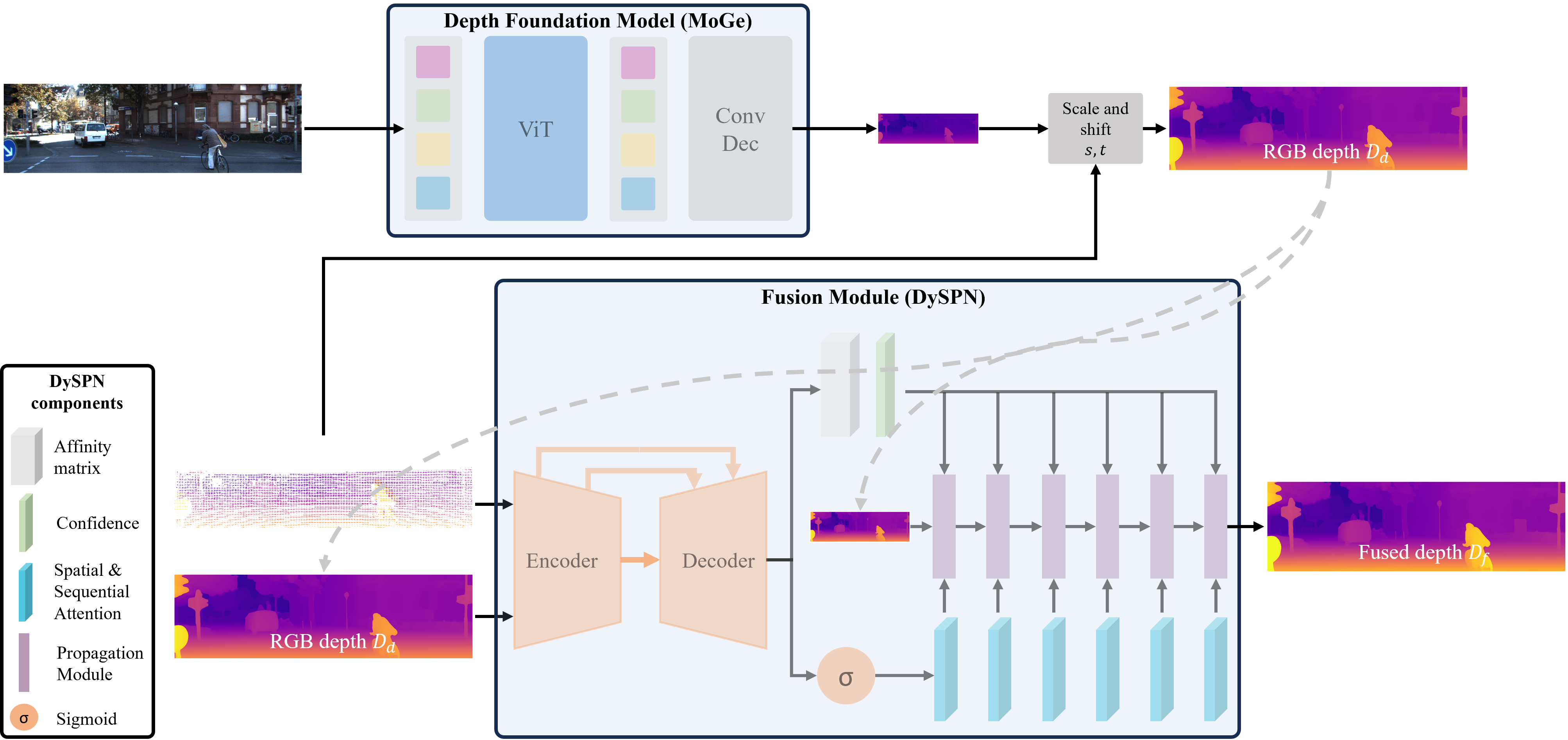}
    \caption{\textbf{Architecture overview.} A frozen monocular depth foundation model predicts a dense, scale-ambiguous depth prior from the input RGB image. After per-frame alignment to raw sparse LiDAR, the aligned prior and sparse LiDAR depth are fused to predict dense metric depth, including beyond LiDAR coverage. In our implementation, we use MoGe~\cite{MoGe2025} as the foundation model and DySPN~\cite{DySPN_2023} as the fusion module.}
    \label{fig:architecture}
\end{figure*}

\subsubsection{Monocular Prior and Per-Frame Alignment}

We first use a frozen monocular depth foundation model to obtain a dense depth prior. Although this prior captures detailed scene geometry, it remains subject to global scale and shift ambiguity as well as residual calibration bias. We therefore align it to the raw LiDAR scan on a per-frame basis, following the point-cloud calibration procedure of MoGe~\cite{MoGe2025}.

Concretely, we back-project the dense prior into a point map $\hat{\mathbf{x}}_{u,v}$ in the camera frame, and project the LiDAR scan into the image to obtain LiDAR points $\mathbf{x}^{\mathrm{L}}_{u,v}$ at valid pixels $(u,v)\in\Omega_s$. We then estimate a global scale $s$ and a longitudinal shift $t_z$ along the optical axis by solving
\begin{equation}
\min_{s,\,t_z}\;\sum_{(u,v)\in\Omega_s}
\frac{1}{z}\left\|\, s\big(\hat{\mathbf{x}}_{u,v} + (0,0,t_z)^\top\big) - \mathbf{x}^{\mathrm{L}}_{u,v}\,\right\|_1,
\label{eq:alignment_objective}
\end{equation}
where $z$ denotes the LiDAR depth at pixel $(u,v)$. This produces an aligned dense depth prior $D_d$ in the camera frame.

\subsubsection{Depth Fusion from Aligned Prior and Sparse LiDAR}

Given the aligned dense prior $D_d$ and the sparse raw LiDAR depth map $D_s$, we predict dense metric depth $D_f$ with a lightweight refinement network that fuses both signals. In our implementation, we use Dynamic Spatial Propagation Network (DySPN)~\cite{DySPN_2023} as the fusion backbone. DySPN iteratively refines an initial depth estimate using learned, iteration-dependent neighbor affinities, and serves here as a representative propagation-based completion module. We initialize DySPN with the aligned prior $D_d$ and use $D_s$ as sparse metric anchors.

Our framework is not tied to DySPN: any module that refines a dense prior using sparse depth anchors can be used in its place. Providing both inputs in the depth domain avoids the need for a separate RGB-based initial prediction and focuses the refinement stage on reconciling two geometric signals: preserving the dense structure of the prior while incorporating reliable sparse metric evidence.

% Under raw sparse supervision, localized spurious LiDAR measurements can be propagated into dense predictions as geometric artifacts. This motivates the robust sparse anchor and high-frequency geometric regularization introduced below.

\subsection{Training Objective}

The fusion module is trained using only the sparse and potentially noisy LiDAR depth $D_s$ together with the aligned dense prior $D_d$. The objective combines a sparse anchor term for metric consistency at valid LiDAR returns with geometric regularization that transfers surface structure from the prior. Two components are standard: a robust sparse anchor loss and a normal-consistency regularizer. The additional high-frequency term applies total variation to the per-pixel cosine discrepancy.

\subsubsection{Robust Sparse Anchor Loss}

Raw single-sweep LiDAR can contain outliers due to mixed pixels, multipath, occlusions, reflections, or hidden points. To reduce sensitivity to large residuals while still anchoring the prediction to most valid returns, we use a pruned $L_1$ loss. We write $\Omega_s$ for the set of pixels with valid LiDAR depth and $N=|\Omega_s|$. Let
\begin{equation}
r_i = \left|D_f(i) - D_s(i)\right|,\quad i\in\Omega_s,
\end{equation}
and let $\delta_q$ be the $q$-percentile of $\{r_i\}_{i\in\Omega_s}$. We keep the inlier set
\begin{equation}
\mathcal{I}_q = \left\{ i\in\Omega_s : r_i \le \delta_q \right\},
\end{equation}
and define
\begin{equation}
\mathcal{L}_{\mathrm{RSA}}(D_f, D_s)
= \frac{1}{|\mathcal{I}_q|}\sum_{i\in\mathcal{I}_q} r_i.
\label{eq:RSA}
\end{equation}
This corresponds to a simple pruned estimator: standard $L_1$ is applied to the majority of LiDAR points, while the largest residuals, which are more likely to be spurious, are discarded.

\subsubsection{Surface Normal Consistency Regularizer}

To preserve the coherent surface structure encoded by the dense prior, we encourage surface normals computed from $D_f$ to agree with those from $D_d$. For each pixel $(u,v)\in\Omega$, we back-project depth to 3D points $\mathbf{x}(u,v)$ in the camera frame and compute unit normals
\begin{equation}
\hat{\mathbf{n}}(u,v) =
\frac{(\partial_u \mathbf{x}) \times (\partial_v \mathbf{x})}
{\|(\partial_u \mathbf{x}) \times (\partial_v \mathbf{x})\|}.
\label{eq:normal_vector}
\end{equation}
We then penalize their angular discrepancy via cosine distance:
\begin{equation}
\mathcal{L}_{\mathbf{n}}(D_f, D_d)
= \frac{1}{|\Omega|}\sum_{(u,v)\in\Omega}
\left(1 - \hat{\mathbf{n}}_f(u,v)\cdot \hat{\mathbf{n}}_d(u,v)\right).
\label{eq:normal_cosine}
\end{equation}
This term regularizes the prediction toward the surface structure of the aligned prior while still allowing metric corrections from sparse LiDAR anchors.

\subsubsection{High-Frequency Normal Consistency}

While $\mathcal{L}_{\mathbf{n}}$ constrains normals globally, it is often insufficient to suppress localized geometric corruption caused by noisy sparse measurements. Under raw sparse supervision, such corruption can arise when spurious LiDAR returns are copied into the dense prediction as local spikes, punctures, or ripples. We therefore introduce a high-frequency normal-consistency loss that penalizes spatial variation in the per-pixel normal discrepancy.

Let
\begin{equation}
\ell_{\mathbf{n}}(u,v) = 1 - \hat{\mathbf{n}}_f(u,v)\cdot \hat{\mathbf{n}}_d(u,v).
\label{eq:normal_cosine_pixel}
\end{equation}
We define
\begin{equation}
\mathcal{L}_{\nabla}(D_f, D_d)
= \frac{1}{|\Omega|}\sum_{(u,v)\in\Omega}
\left(\left|\partial_u \ell_{\mathbf{n}}(u,v)\right|
+ \left|\partial_v \ell_{\mathbf{n}}(u,v)\right|\right),
\label{eq:grad_normal_cosine}
\end{equation}
where $\partial_u$ and $\partial_v$ are finite differences in image space.

Intuitively, $\mathcal{L}_{\nabla}$ allows low-frequency deviations from the prior, such as gradual scale or slope corrections, while penalizing short-scale inconsistencies much more strongly. As a result, it suppresses localized artifacts introduced by noisy LiDAR measurements without preventing the prediction from adapting to the large-scale geometry supported by sparse LiDAR anchors. 
% A frequency-domain analysis is provided in Appendix~\ref{app:fourier_hfnc}: differentiating the per-pixel normal discrepancy amplifies higher spatial frequencies, so $\mathcal{L}_{\nabla}$ primarily penalizes localized deviations while leaving low-frequency structure to be determined by the sparse anchor term.

\paragraph{Why the gradient normal-consistency loss emphasizes local deviations.}
To make the effect of $\mathcal{L}_{\nabla}$ more explicit, consider the per-pixel normal-consistency term in Eq.~\ref{eq:normal_cosine_pixel} and fix one image row $v$. Writing $\ell_{\mathbf{n}}^u \equiv \ell_{\mathbf{n}}(u,v)$ as a Fourier series along the horizontal image coordinate $u$,
\begin{equation}
    \ell_{\mathbf{n}}^u
    =
    \frac{a_0}{2}
    + \sum_{k\geq 1}
    \left(
        a_k \cos\!\left(\frac{2\pi k}{W}u\right)
        +
        b_k \sin\!\left(\frac{2\pi k}{W}u\right)
    \right),
\end{equation}
where $W$ denotes the image width. Differentiating with respect to $u$ gives
\begin{equation}
    \partial_u \ell_{\mathbf{n}}^u
    =
    \frac{2\pi}{W}
    \sum_{k\geq 1}
    k\left(
        b_k \cos\!\left(\frac{2\pi k}{W}u\right)
        -
        a_k \sin\!\left(\frac{2\pi k}{W}u\right)
    \right).
\end{equation}
Hence, the contribution of Fourier mode $k$ is scaled by an additional factor of $k$. Higher-frequency components, corresponding to shorter spatial length scales, are therefore amplified more strongly in the derivative than low-frequency components. 

Finally, we write the full loss as
\begin{equation}
\mathcal{L}
= \lambda_{\mathrm{RSA}}\,\mathcal{L}_{\mathrm{RSA}}(D_f, D_s)
+ \lambda_{\mathbf{n}}\,\mathcal{L}_{\mathbf{n}}(D_f, D_d)
+ \lambda_{\nabla}\,\mathcal{L}_{\nabla}(D_f, D_d),
\label{eq:loss_total}
\end{equation}
with weights $\lambda_{\mathrm{RSA}},\lambda_{\mathbf{n}},\lambda_{\nabla}$.

\section{Experiments}
\label{sec:experiments}

We evaluate the proposed framework in the intended sparse-only, single-frame supervision setting, with particular emphasis on extrapolation beyond LiDAR coverage. Our experiments are centered on KITTI, where we report both conventional depth-completion metrics and results under a controlled beyond-coverage protocol. We further use NYUv2 as a clean sparse-only testbed and nuScenes for qualitative robustness analysis in challenging outdoor scenes.

\subsection{Experimental Protocol and Implementation Details}
\label{subsec:exp_setup}

Unless noted otherwise, training uses only raw sparse depth as supervision, dense or semi-dense depth is used only for validation and evaluation. On KITTI, this means training from raw single-sweep velodyne measurements even though standard evaluation is performed against the semi-dense benchmark depth.

\subsubsection{Datasets and evaluation regimes.}
We use three datasets with complementary roles. \textbf{KITTI} is the main benchmark and supports both conventional depth-completion evaluation and our controlled beyond-coverage protocol. \textbf{NYUv2} provides dense and comparatively clean indoor depth, which allows controlled sparse-only experiments without the additional complications of outdoor LiDAR noise and semi-dense benchmark targets. \textbf{nuScenes} is used for qualitative robustness analysis in challenging outdoor scenes with adverse conditions and noisy automotive LiDAR.

\subsubsection{KITTI beyond-coverage protocol.}
For the extrapolation protocol, RGB and depth frames are resized to $336\times1120$. During training, raw LiDAR depth is restricted to a horizontal band in the third quarter of the image and is used both as network input and sparse supervision. Evaluation is then performed on the complementary image region, which contains no direct LiDAR support. This protocol isolates beyond-coverage prediction from interpolation between nearby sparse measurements.

\subsubsection{Baselines.}
We compare against a prior-only baseline obtained by aligning the frozen monocular depth prior without fusion, OMNI-DC~\cite{zuo2025omni} as a foundation depth completion model, and against trained completion baselines including DySPN~\cite{DySPN_2023} and DMD3C~\cite{DMD3c_2025}. For DMD3C~\cite{DMD3c_2025}, we report both the released pretrained model and a variant retrained under our sparse-only protocol.

\subsubsection{Implementation details.}
We implement the model in PyTorch. The fusion backbone is trained from scratch, while the monocular foundation model remains frozen. Unless stated otherwise, we optimize with Adam using a learning rate of $4\times10^{-4}$, batch size $4$, and loss weights $\lambda_{\mathrm{RSA}}=1$, $\lambda_{\mathbf{n}}=0.25$, and $\lambda_{\nabla}=0.5$. For RSA, we trim the largest $4\%$ of sparse-depth residuals. Since we use a frozen MoGe~\cite{MoGe2025} as the prior depth, the full fusion pipeline is trained for only 6 epochs, which on KITTI takes less than two hours on 8 A100 GPUs. At test time inference takes around $\sim80 ms$ per frame on KITTI on a single A100.

\subsection{Extrapolation Beyond LiDAR Coverage on KITTI}
\label{subsec:kitti_extrap}

\begin{figure}
    \centering
    \includegraphics[width=1\linewidth]{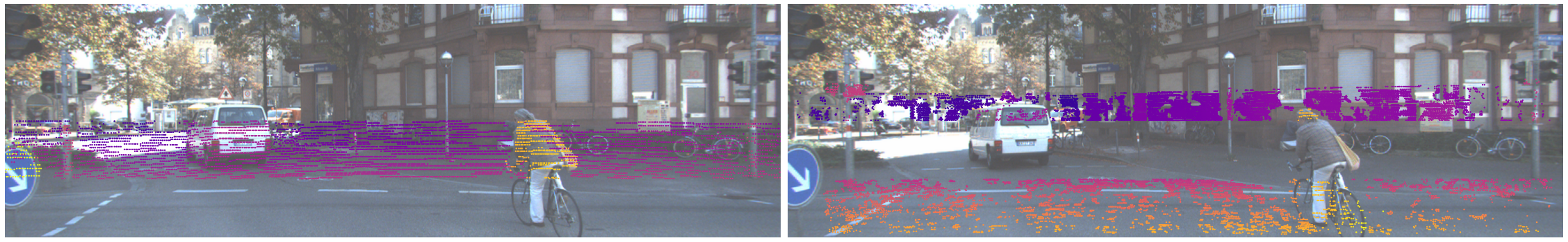}
    \caption{\textbf{KITTI beyond-coverage protocol.} Left: cropped raw Velodyne depth used as input and sparse supervision during training. Right: complementary region used for evaluation only. The two regions are spatially disjoint, so performance in the evaluation region reflects extrapolation beyond direct LiDAR support.}
    \label{fig:kittiextra}
\end{figure}

KITTI~\cite{KITTI_improved_GT} is our main benchmark for evaluating extrapolation beyond LiDAR coverage. Under the controlled protocol described above, LiDAR support is deliberately withheld from the evaluation region: only a horizontal band of raw LiDAR depth is retained during training, while performance is measured on the complementary region (Fig.~\ref{fig:kittiextra}). The task therefore requires prediction in regions with little or no direct LiDAR support rather than interpolation between nearby sparse observations.

\subsubsection{Matched baselines.}
The prior-only baseline isolates the extent to which beyond-coverage performance is derived from the aligned monocular prior alone. DMD3C~\cite{DMD3c_2025} provides a strong learned reference point as it also distills knowledge from a monocular foundation prior, enabling a direct comparison between prior fusion and distillation-based transfer under the same evaluation protocol.

\begin{table}[t]
\centering
% 1. Manually set the font size to match the other tables
\footnotesize 
\renewcommand{\arraystretch}{1.2} 
% 2. Keep padding tight to ensure the text fits without the resizebox
\setlength{\tabcolsep}{3pt}

% (Notice \resizebox is entirely removed here as well)
\begin{tabular}{lcccc}
\toprule
 & \multicolumn{2}{c}{KITTI Extrap.} & \multicolumn{2}{c}{KITTI} \\
\cmidrule(lr){2-3} \cmidrule(lr){4-5} 
Method & RMSE $\downarrow$ & Abs. Rel. $\downarrow$ & RMSE $\downarrow$ & Abs. Rel. $\downarrow$ \\
 & (mm) & (\%) & (mm) & (\%) \\
\midrule
OMNI-DC~$\dagger$~\cite{zuo2025omni} & 3973 & 12.0 & \textbf{1224} & \textbf{1.5} \\
DMD3C~$\dagger$~\cite{DMD3c_2025} & 11220 & 21.1 & 1329 & 2.2 \\
DMD3C~\cite{DMD3c_2025} & 7314 & 12.4 & 2135 & 4.2 \\
DySPN & 16650 & 65.7 & 16092 & 64.6 \\
\midrule
DepthAnything (aligned) & 4491 & 9.5 & 4197 & 8.9 \\
Ours (DepthAnything) & 3728 & 6.8 & 1886 & 2.2 \\
\midrule
MoGe (aligned) & 3652 & 9.7 & 2831 & 5.6 \\
\textbf{Ours (MoGe)} & \textbf{2978} & \textbf{6.7} & 1619 & 2.1 \\
\bottomrule
\end{tabular}
\caption{Results on standard KITTI depth completion and on the controlled beyond-coverage extrapolation protocol. The bottom sections isolate the effect of our fusion strategy against prior-only baselines. $\dagger$ denotes models pretrained on other datasets but evaluated zero-shot on KITTI. ``Aligned'' indicates per-frame metric alignment of the frozen monocular prior to sparse LiDAR anchors without our fusion refinement.}
\label{tab:kitti_results}
\end{table}

\subsubsection{Results.}
\begin{figure}[t]
    \centering
    \includegraphics[width=\linewidth]{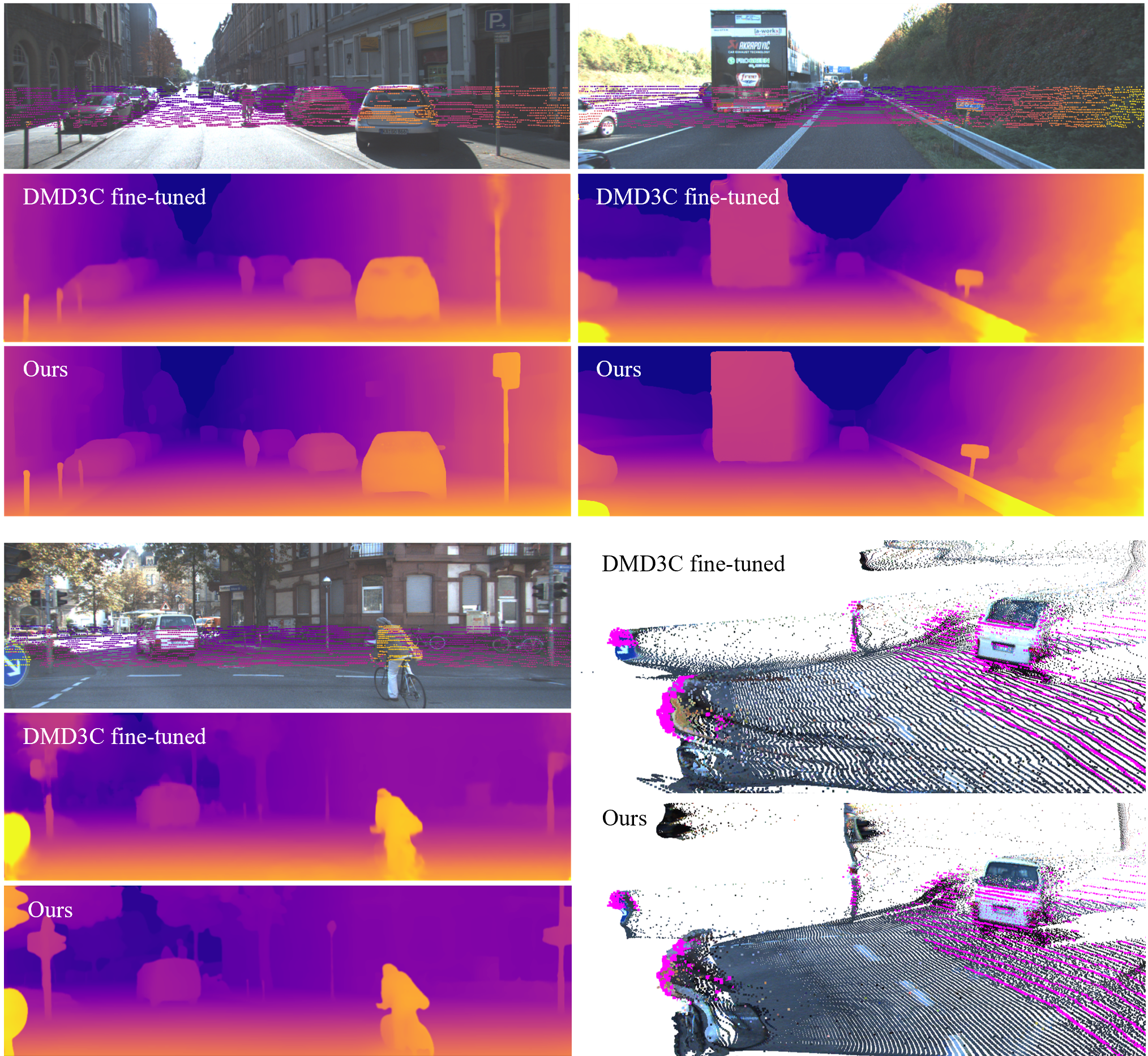}
    % \caption{\textbf{Qualitative results on the KITTI beyond-coverage protocol.} For each example, we show the input RGB image with the cropped raw Velodyne band used as input and sparse supervision, followed by the predictions of DMD3C and our method. For the bottom example, a corresponding colored point cloud is shown on the right. Our formulation yields more coherent surfaces, preserves thin structures more faithfully, and reconstructs more plausible geometry in regions beyond direct LiDAR support.}

    \caption{\textbf{Qualitative results on the KITTI beyond-coverage protocol.} We show the RGB input with the cropped Velodyne band (sparse supervision), followed by DMD3C and our predictions. Our approach yields more coherent surfaces, preserves thin structures, and reconstructs plausible geometry beyond direct LiDAR support.}
    \label{fig:kitti_extra_merged}
\end{figure}

Table~\ref{tab:kitti_results} shows that the proposed formulation substantially improves beyond-coverage prediction over both the aligned prior alone and the learned baselines. In particular, fusion with sparse LiDAR anchors clearly improves over the prior-only baseline, indicating that the gains cannot be explained by the monocular prior alone. Under this protocol, our method also outperforms both the released pretrained DMD3C checkpoint and the sparse-only retrained variant by a large margin, suggesting that direct fusion of an aligned frozen prior is especially effective when the evaluation region contains no LiDAR support, as shown in~\Cref{fig:kitti_extra_merged}.

Qualitatively, the proposed framework extrapolates thin structures more coherently across the masked region, preserves smoother surfaces, and recovers plausible geometry for objects disconnected from direct LiDAR support (Fig.~\ref{fig:kitti_extra_merged}). DMD3C degrades more strongly outside the LiDAR-supported band, particularly on thin structures and in the reconstructed point clouds.

\subsection{Sparse-Only Depth Completion on KITTI}
\label{subsec:kitti_standard}

We also evaluate the proposed framework on the standard KITTI depth completion benchmark, where the reported metrics are dominated by LiDAR-supported regions. In this regime, our method remains competitive under sparse-only supervision (Table~\ref{tab:kitti_results}). While it does not outperform the released pretrained DMD3C checkpoint on all metrics, it improves over the sparse-only retrained DMD3C model and produces accurate, geometrically coherent reconstructions without dense supervision, LiDAR accumulation, or dataset-specific densification.

This comparison complements the beyond-coverage results with the conventional in-coverage setting. Together, the two evaluations show that the proposed formulation remains effective in standard LiDAR-covered completion, while its main advantage emerges in the beyond-coverage regime. Qualitatively, our model is also less sensitive to localized sparse artifacts in LiDAR-covered regions, consistent with the role of the robust sparse anchor and high-frequency normal-consistency losses (Fig.~\ref{fig:kitti_standard_qual}).

\subsection{Controlled Sparse-Only Completion on NYUv2}
\label{subsec:nyuv2}

We use NYUv2 as a controlled sparse-only testbed. Its dense, comparatively clean indoor depth allows us to simulate different sparsity patterns while avoiding the additional ambiguities of outdoor LiDAR noise and semi-dense benchmark targets. As shown in~\Cref{tab:nyu_500}, given 500 depth points as input, our method remains competitive with DMD3C~\cite{DMD3c_2025} despite using only the sparse points for training, whereas DMD3C~\cite{DMD3c_2025} is supervised by full dense ground truth. This highlights the effectiveness of our sparse-only formulation even under restricted supervision. Additional results across sampling densities and sparsity patterns are reported in the appendix. Performance improves with denser sampling and remains stable even under structured LiDAR-like coverage gaps.

\begin{table}[t]
\centering
% 1. Manually set the font size to match the scaled-down KITTI tables
\footnotesize 
\renewcommand{\arraystretch}{1.2} 
% 2. We can relax the column padding slightly since we aren't forcing a fit
\setlength{\tabcolsep}{4pt}

% (Notice \resizebox is entirely removed here)
\begin{tabular}{lccccc}
\toprule
Method & Abs Rel $\downarrow$ & RMSE $\downarrow$ & $\delta_1 \uparrow$ & $\delta_2 \uparrow$ & $\delta_3 \uparrow$ \\
 & & (mm) & & & \\
\midrule
DMD3C~\cite{DMD3c_2025} & \textbf{0.028} & \textbf{1141} & \textbf{0.984} & \textbf{0.996} & \textbf{0.998} \\
Ours & 0.030 & 1166 & 0.947 & 0.981 & 0.991 \\
\bottomrule
\end{tabular}
\caption{Comparison on NYUv2~\cite{NYU_dataset} with 500 sparse depth samples as input. DMD3C is trained with full dense ground-truth depth supervision, whereas our framework uses only the 500 sparse samples as supervision.}
\label{tab:nyu_500}
\end{table}

\subsection{Ablations and Robustness Analysis}
\label{subsec:ablations}

\begin{figure}[t]
    \centering
    \includegraphics[width=\linewidth]{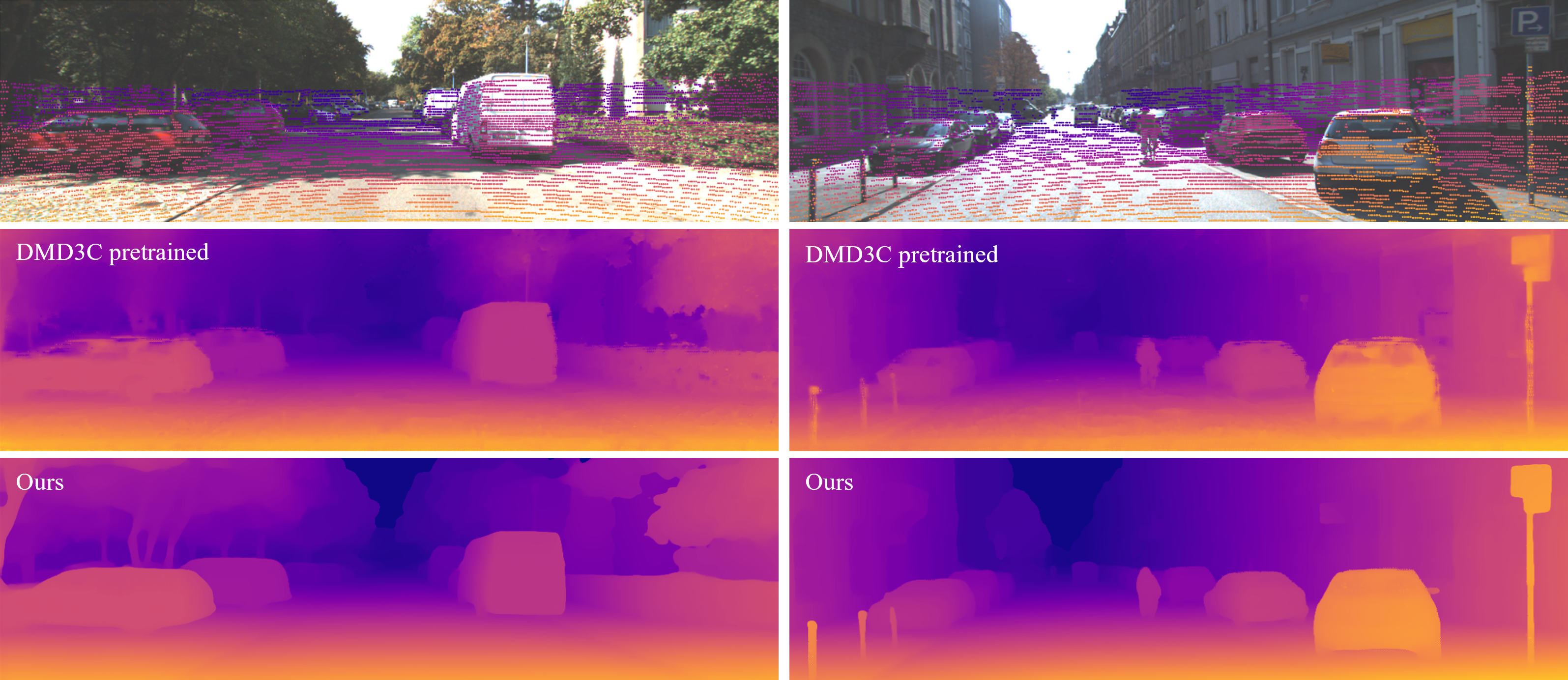}
    \caption{\textbf{Qualitative results on KITTI-DC~\cite{KITTI_improved_GT}.} The first row shows the full raw Velodyne input provided to both the pretrained DMD3C in the second row and our method in the third row. The proposed framework produces more geometrically coherent reconstructions while reducing localized artifacts.}

    % \caption{\textbf{Qualitative results on the KITTI benchmark.} Given the raw Velodyne input (top), our method (bottom) yields more geometrically coherent reconstructions with fewer localized artifacts compared to pretrained DMD3C (middle).}
    \label{fig:kitti_standard_qual}
\end{figure}

\subsubsection{Component Ablations Under Matched Protocols}

\begin{figure*}[t]
    \centering
    \includegraphics[width=1\textwidth]{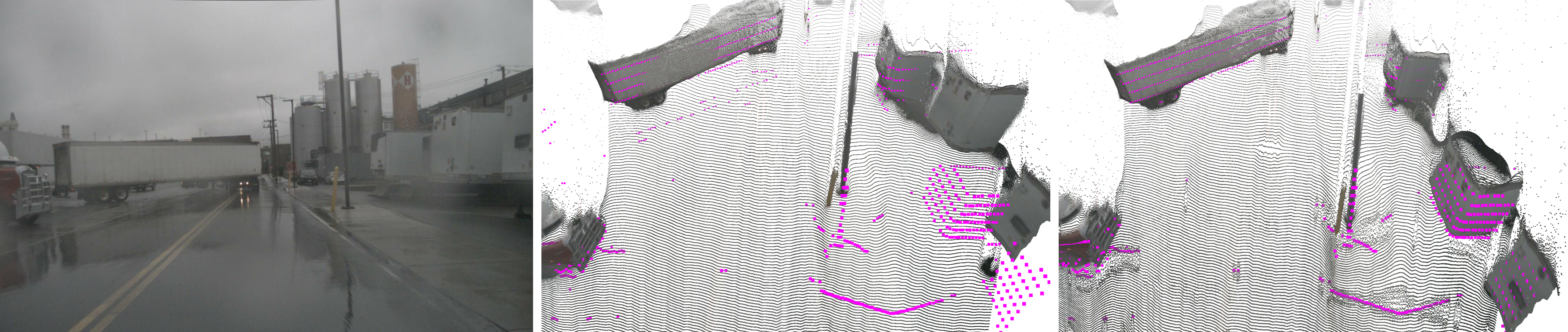}
    \caption{\textbf{Qualitative robustness analysis on a challenging nuScenes scene.} \textit{Left}: input RGB image with reflective wet-road surfaces and large weakly textured regions. \textit{Center}: point cloud reconstructed from the monocular prior together with raw LiDAR, showing both large-scale prior inconsistencies and local LiDAR outliers. \textit{Right}: point cloud produced by the proposed framework together with filtered LiDAR points. The fused reconstruction follows sparse LiDAR geometry at large scales while suppressing short-scale artifacts, including below-ground reflections on the road surface.}
    \label{fig:nusceneschallenge}
\end{figure*}

Table~\ref{tab:foundationfusion_ablation} summarizes component ablations on both standard KITTI and the beyond-coverage protocol. Removing either the aligned dense prior (\textit{w/o depth input}) or the fusion stage (\textit{w/o completion}) causes a large degradation, especially in the extrapolation regime, showing that strong beyond-coverage performance requires both dense monocular structure and explicit fusion with sparse LiDAR anchors. Standard normal regularization also provides steady improvements. The global metrics do not establish a benefit from the high-frequency term in the extrapolation region: removing it slightly improves RMSE and Abs. Rel. (2,949\,mm and 6.2\% versus 2,978\,mm and 6.7\%), while worsening the standard KITTI metrics (1,649\,mm and 2.4\% versus 1,619\,mm and 2.1\%). Its effect is illustrated qualitatively in~\Cref{fig:nusceneschallenge}, region-specific quantitative validation remains future work.

The behavior of RSA depends on the evaluation mask. On the extrapolation-only region, removing RSA can slightly improve the reported metrics because this mask contains no direct LiDAR support and therefore does not expose the sparse outliers that RSA is designed to suppress. On the full semi-dense validation depth, however, RSA is clearly beneficial ($\mathrm{RMSE}=2398$\,mm, $\mathrm{AbsRel}=4.5\%$ versus $2889$\,mm and $8.9\%$), confirming its value when evaluation includes LiDAR-supported regions.

\begin{table}[t]
\centering
% 1. Manually set the font size to match the NYUv2 table
\footnotesize 
\renewcommand{\arraystretch}{1.2} 
% 2. Keep padding tight to ensure the text fits without the resizebox
\setlength{\tabcolsep}{3pt}

% (Notice \resizebox is removed here as well)
\begin{tabular}{lcccc}
\toprule
 & \multicolumn{2}{c}{KITTI} & \multicolumn{2}{c}{KITTI Extrap.} \\
\cmidrule(lr){2-3} \cmidrule(lr){4-5}
Variant & RMSE $\downarrow$ & Abs. Rel. $\downarrow$ & RMSE $\downarrow$ & Abs. Rel. $\downarrow$ \\
 & (mm) & (\%) & (mm) & (\%) \\
\midrule
w/o normal consist. & 1992 & 2.4 & 3013 & 6.9 \\
w/o HF norm. consist. & 1649 & 2.4 & 2949 & 6.2 \\
w/o RSA & 1782 & 2.5 & \textbf{2938} & \textbf{6.1} \\
w/o depth input & 4739 & 3.3 & 8807 & 16.9 \\
w/o completion & 2831 & 5.6 & 3652 & 9.7 \\
\textbf{Full model} & \textbf{1619} & \textbf{2.1} & 2978 & 6.7 \\
\bottomrule
\end{tabular}
\caption{Component ablations on standard KITTI and on the beyond-coverage protocol. The \textit{w/o depth input} variant removes the aligned dense prior from the fusion network, while \textit{w/o completion} corresponds to the aligned monocular prior without refinement. The extrapolation metrics are evaluated only on the complementary beyond-coverage region, which explains why removing RSA can slightly improve this mask-specific score despite hurting performance on the full semi-dense validation set.}
\label{tab:foundationfusion_ablation}
\end{table}
\subsubsection{Effect of the High-Frequency Normal-Consistency Loss}

Although its impact is not always reflected by aggregate depth-completion metrics, the high-frequency normal-consistency loss noticeably changes depth-completion behavior. Localized artifacts often occur in regions that are sparsely evaluated or filtered in commonly used reference depth (e.g., near depth discontinuities or near invalid/occluded pixels), so propagating spurious structure may contribute little to reported averages. Figure~\ref{fig:kitti_standard_qual} illustrates this phenomenon on standard KITTI: under sparse-only supervision, localized noise can remain visible in dense predictions even when benchmark metrics are strong, whereas our formulation yields cleaner geometry in these regions. We use the released pretrained DMD3C checkpoint for this comparison because it matches our sparse-only protocol, while DMD3C’s final model is trained with dense ground truth and does not exhibit these artifacts~\cite{DMD3c_2025}. This highlights that such qualitative effects can be sensitive to the supervision regime and may be only weakly reflected by aggregate scores.

We further illustrate the role of the high-frequency loss on nuScenes, which contains adverse conditions such as wet roads, weakly textured surfaces, and noisy automotive LiDAR but does not provide cleaned dense depth ground truth. In the example of Fig.~\ref{fig:nusceneschallenge}, wet-road reflections produce below-ground LiDAR outliers that can puncture the reconstructed surface. In this example, reflected sparse outliers appear more readily propagated into the fused prediction without the term, while the corresponding short-scale deviations are visually reduced with it. This comparison is qualitative and does not provide a quantitative measure of artifact suppression.

% Although its effect on aggregate metrics is modest, the high-frequency normal-consistency loss changes the qualitative behavior of the model in challenging scenes. We illustrate this on nuScenes, which contains adverse conditions such as wet roads, weakly textured surfaces, and noisy automotive LiDAR, but does not provide cleaned dense depth ground truth. In the example of Fig.~\ref{fig:nusceneschallenge}, the monocular prior exhibits large-scale inconsistencies, while wet-road reflections produce below-ground LiDAR outliers that can puncture the reconstructed surface. The high-frequency loss is particularly important for the latter case: without it, reflected sparse outliers are more readily copied into the fused prediction, whereas with it, these short-scale deviations are suppressed more effectively.

\subsubsection{Robustness to Sparse LiDAR Outliers}

We additionally verify the effect of RSA under synthetic sparse outlier corruption, a critical requirement for production-ready raw LiDAR fusion. Because standard global metrics often average out severe localized artifacts, we demonstrate in the appendix that RSA effectively isolates and reduces the propagation of corrupted sparse measurements into the reconstructed geometry, yielding structurally robust depth maps where baselines fail.

% \subsubsection{Effect of the Monocular Prior}

% To isolate our fusion strategy's contribution from the foundation model itself, we evaluate our pipeline using both MoGe and DepthAnything. We compare against prior-only baselines that are already metrically aligned per-frame to the sparse LiDAR anchors to resolve scale and shift. 

% As shown in \cref{tab:kitti_results}, our method yields substantial gains regardless of the chosen prior. Applying our fusion to DepthAnything reduces error over its aligned baseline by 17.0\% RMSE (30.9\% Abs.\ Rel.) on KITTI extrapolation, and by 55.1\% RMSE on standard KITTI. Because the baselines already incorporate sparse anchors globally, these improvements confirm that our module actively corrects local prior geometry rather than merely inheriting the foundation model's capabilities. While MoGe remains the strongest base, our formulation successfully adapts varying priors for both standard completion and beyond-coverage extrapolation.

\section{Conclusion}
\label{sec:conclusion}

We studied depth completion beyond LiDAR coverage in the sparse-only supervision regime, where dense metric depth must be inferred from a single RGB image and raw single-sweep LiDAR without dense labels, LiDAR accumulation, or pseudo-LiDAR generation. We showed that a simple fusion formulation based on a frozen monocular depth prior, lightweight per-frame alignment, and sparse LiDAR refinement performs strongly in this setting.

A central challenge in this regime is robustness to localized artifacts induced by raw sparse measurements. We introduce a high-frequency normal-consistency loss that penalizes spatial variation in scalar angular discrepancy; its effect on localized artifacts is demonstrated qualitatively, but is not established by the reported global metrics.

Beyond the method itself, we argued that conventional depth-completion evaluation under-emphasizes extrapolation and does not fully reflect robustness to sparse measurement artifacts. We therefore introduced a controlled protocol that explicitly separates LiDAR-supported and beyond-coverage regions. Under this protocol, the proposed framework achieves strong extrapolation performance while remaining competitive on standard KITTI depth completion.

Overall, our results suggest that sparse-only fusion with frozen monocular priors is a practical and effective operating point for real camera--LiDAR systems. We hope that the proposed formulation and evaluation protocol help make extrapolation beyond coverage and sparse-only supervision more explicit targets in future depth-completion research.

{
    \small
    \bibliographystyle{ieeenat_fullname}
    \bibliography{main}
}

\clearpage
\appendix
% Supplementary sections included by main.tex after the references.

\section{Additional NYUv2 Results}
\label{sec:appendix_nyuv2_additional}

Table~\ref{tab:nyu_suptype_moge} reports additional NYUv2 results under different sparsity patterns and sampling densities. Across settings, the proposed formulation yields stable and accurate reconstructions under sparse-only supervision. Performance improves with denser sampling, while the LiDAR-like patterns confirm that the framework remains effective even under persistent structured coverage gaps. The first row also shows that replacing the geometric regularization with direct $L_1$ depth supervision alone leads to substantially weaker results.

\begin{table}[ht]
\footnotesize
\centering
\renewcommand{\arraystretch}{1.3} 
\setlength{\tabcolsep}{5pt}
\begin{tabular}{lcc}
\toprule
Setting & Abs. Rel. $\downarrow$ & RMSE (mm) $\downarrow$ \\
\midrule
Rand., 5k pts, $L_1$ only & 0.0520 & 2502 \\
\midrule
LiDAR-like, 16 beams & 0.0517 & 1920 \\
LiDAR-like, 64 beams & 0.0214 & 1022 \\
Rand., 500 pts & 0.0303 & 1166 \\
Rand., 5k pts & 0.0270 & 900 \\
\bottomrule
\end{tabular}
\caption{NYUv2 results under sparse-only supervision for different sparsity patterns and sampling densities. The first row replaces the geometric regularization with direct $L_1$ depth supervision only. The remaining rows use the proposed formulation without pruning the $L_1$ loss, which is not needed in this noise-free sparsification setting. Performance improves with denser sampling, while the LiDAR-like patterns show that the framework remains effective even under persistent structured coverage gaps.}
\label{tab:nyu_suptype_moge}
\end{table}

\section{Robustness to Sparse LiDAR Outliers}
\label{sec:appendix_sparse_outliers}

To further isolate the role of the robust sparse anchor (RSA) loss, we inject synthetic large-depth outliers into the sparse input depth within a selected rectangular image region. Concretely, we sample pixel locations inside the box and overwrite small local patches with far-range depth values, producing compact clusters of spurious far-range returns similar to multipath or reflection artifacts. This creates a controlled stress test for robustness to sparse outliers.

\begin{figure*}[t]
    \centering
    \includegraphics[width=\textwidth]{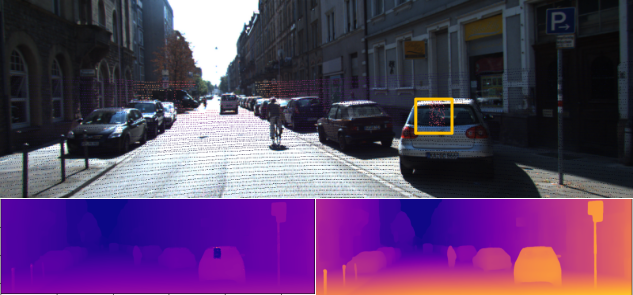}
    \caption{Effect of the robust sparse anchor (RSA) loss under synthetic sparse outlier corruption. \textbf{\textit{Top}}: RGB overlay with injected large-depth outliers highlighted by a bounding box. \textbf{\textit{Bottom left}}: model trained without RSA; the corrupted sparse points are propagated into the reconstruction and distort the local geometry. \textbf{\textit{Bottom right}}: full model with RSA; the same outliers are rejected and the local surface remains coherent. RSA trims the largest sparse depth residuals during training, encouraging the network to ignore spurious LiDAR measurements.}
    \label{fig:appendix_os_ablation_grid}
\end{figure*}

Figure~\ref{fig:appendix_os_ablation_grid} shows that a model trained without RSA propagates these corrupted sparse measurements into the surrounding reconstruction, whereas the full model rejects them and preserves the local surface geometry. In the example shown here, the reconstruction remains visually clean even after injecting $N=200$ outlier points, with identical RMSE ($0.94$\,m). While this synthetic perturbation does not replace a dedicated real-world outlier benchmark, it supports the intended role of RSA as a simple robust estimator under raw sparse LiDAR supervision.

% ====================================================================
% DOUBLE-COLUMN FIGURE
% ====================================================================
% ====================================================================

\end{document}